\documentclass[11pt,a4paper]{article}
\usepackage[hyperref]{emnlp2020}
\usepackage{times}
\usepackage{textcomp}
\usepackage{multirow}
\usepackage{amsmath}
\usepackage{latexsym}
\usepackage{graphicx}
\usepackage{multirow,bigdelim,dcolumn,booktabs}
\usepackage{tikz-dependency}

\usepackage{color, colortbl}
\usepackage{amssymb}

\usepackage{microtype}

\aclfinalcopy 

\newcommand{\LT}[0]{{\sc LaserTagger}}
\newcommand{\GT}[0]{{\sc Felix}}

\newcommand{\insertion}[0]{\textit{insertion}}
\newcommand{\tagging}[0]{\textit{tagging}}
\newcommand{\tagger}[0]{\textit{tagger}}
\newcommand{\INSERTION}[0]{{\sc FelixInsert}} 
\newcommand{\LP}[0]{{\sc FelixPoint}} 
\newcommand{\seqtoseqbert}[0]{{\sc seq2seq}\textsubscript{BERT}}
\newcommand{\bertobert}[0]{{\sc Bert2Bert}}
\newcommand{\LevT}[0]{{\sc LevT}}
\newcommand{\tags}[1]{\texttt{#1}}

\newcommand{\bv}[1]{\mathbf{#1}}

\title{Felix: Flexible Text Editing Through Tagging and Insertion}

\author{Jonathan Mallinson\thanks{\hspace{0.15cm} Equal contribution.}\hspace{0.05cm} \thanks{\hspace{0.15cm} Work completed during internship at Google Research.}\\
 University of Edinburgh \\
  \texttt{J.Mallinson@ed.ac.uk}   \\
  \And
  Aliaksei Severyn\footnotemark[1] \\
  Google Research \\
    \\
  \And
  Eric Malmi \\ 
  Google Research \\
  \texttt{\{severyn,emalmi,ggarrido\}@google.com} \\
  \And
  Guillermo Garrido \\ 
  Google Research
  }

\date{}

\begin{document}
\maketitle
\begin{abstract}
We present \GT{} -- a flexible text-editing approach for generation, designed to derive the maximum benefit from the ideas of decoding with bi-directional contexts and self-supervised pre-training. In contrast to conventional sequence-to-sequence (seq2seq) models, \GT{} is efficient in low-resource settings and fast at inference time, while being capable of modeling flexible input-output transformations. We achieve this by decomposing the text-editing task into two sub-tasks: \tagging{} to decide on the subset of input tokens and their order in the output text and \insertion{} to in-fill the missing tokens in the output not present in the input. The \tagging{} model employs a novel Pointer mechanism, while the \insertion{} model is based on a Masked Language Model. Both of these models are chosen to be non-autoregressive to guarantee faster inference. \GT{} performs favourably when compared to recent text-editing methods and strong seq2seq baselines when evaluated on four NLG tasks: Sentence Fusion, Machine Translation Automatic Post-Editing, Summarization, and Text Simplification.
\end{abstract}

\section{Introduction}

\begin{figure}[tb]
\centering
\includegraphics[width=\linewidth]{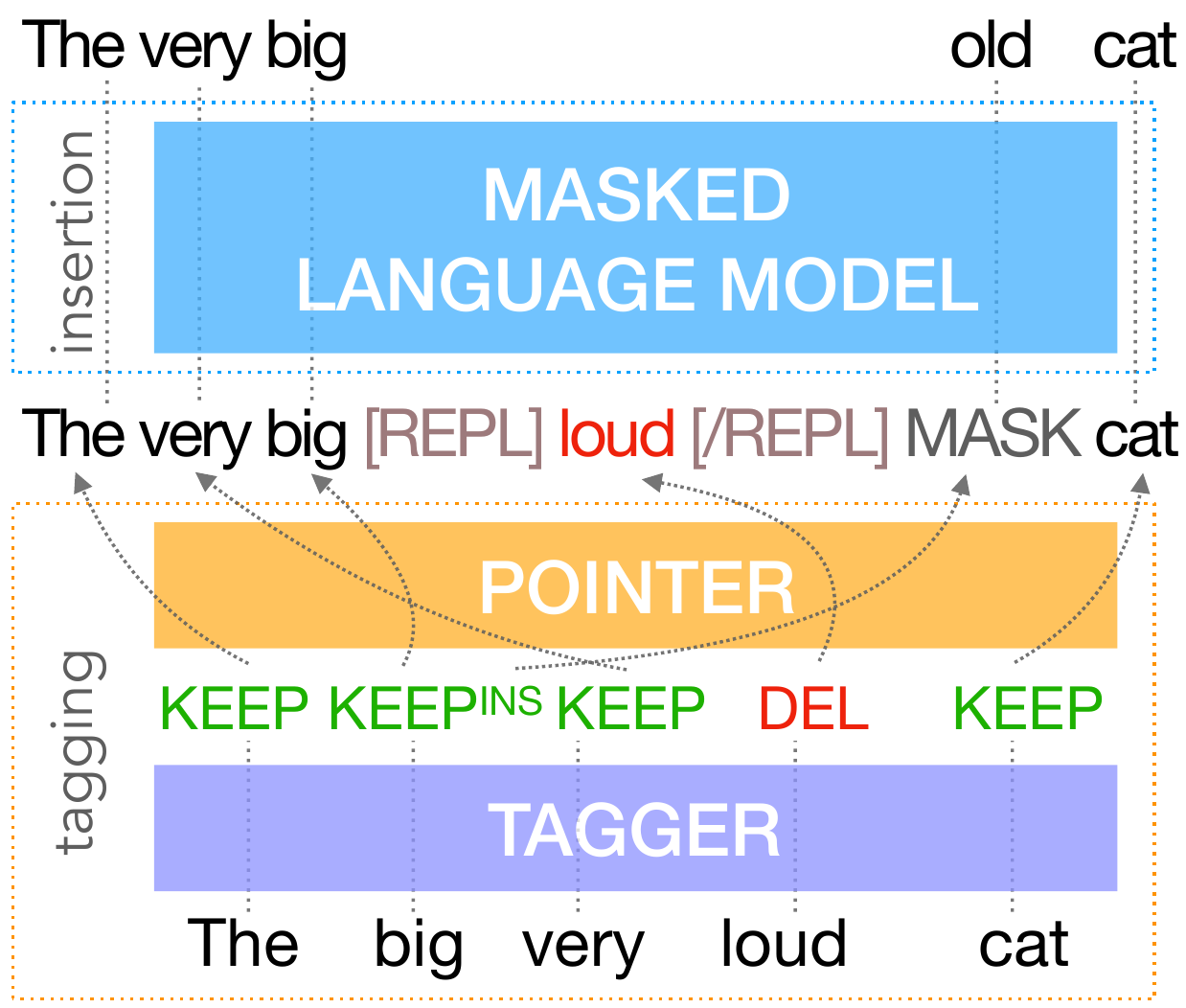}
\caption{\GT{} transforms the source \textit{``The big very loud cat"} into the target text \textit{``The very big old cat"}.}
\label{fig:gutentag}
\end{figure}

The idea of text in-filling when coupled with the self-supervised pre-training of deep Transformer networks on large text corpora have dramatically changed the landscape in Natural Language Understanding. BERT~\cite{Bert} and its successive refinements RoBERTa~\cite{Roberta}, ALBERT~\cite{Albert} implement this recipe and have significantly pushed the state-of-the-art on multiple NLU benchmarks such as GLUE~\cite{GLUE} and SQuAD~\cite{SQUAD}.

More recently, the idea of using masked or in-filling style objective for model pretraining has also been applied to sequence-to-sequence tasks and has significantly pushed the state-of-the-art on a number of text generation tasks, e.g, KERMIT~\cite{KERMIT}, MASS~\cite{MASS}, Bert2Bert~\cite{Bert2Bert}, BART~\cite{BART} and T5~\cite{T5}.

While sequence-to-sequence frameworks offer a generic tool for modeling almost any kind of text-to-text transduction, there are still many real-world tasks where generating target texts completely from scratch---as it is done with the seq2seq approaches---can be unnecessary. This is especially true for monolingual settings where input and output texts have relatively high degrees of overlap. In such cases a natural approach is to cast the task of conditional text generation into a text-editing task, where the model learns to reconstruct target texts by applying a set of edit operations to the inputs. Typically, the set of edit operations is fixed and pre-defined ahead of time, which on one hand limits the flexibility of the model to reconstruct arbitrary output texts from their inputs, but on the other leads to higher sample-efficiency as the limited set of allowed operations significantly reduces the search space. Based on this observation, text-editing approaches have recently re-gained significant interest~\cite{LevenshteinTransformer,EditNTS,awasthi2019parallel,LaserTagger}.

In this paper we present a novel text-editing framework, \GT{}, which is heavily inspired by the ideas of bi-directional decoding (slot in-filling) and self-supervised pre-training. In particular, we have designed \GT{} with the following requirements in mind: 

\textbf{Sample efficiency.}
Training a high precision text generation model typically requires large amounts of high-quality supervised data. Self-supervised techniques based on text in-filling  have been shown to provide a crucial advantage in low-resource settings.
Hence, we focus on approaches able to benefit from already existing pre-trained language models such as BERT, where the final model is directly fine-tuned on the down-stream task.

\textbf{Fast inference time}. 
Achieving low latencies when serving text generation models typically requires specialized hardware and finding a trade-off between model size and accuracy. One of the major reasons for slow inference times is that text generation models typically employ an autoregressive decoder, i.e., output texts are generated in a sequential non-parallel fashion. To ensure faster inference times we opt for keeping \GT{} fully \textit{non-autoregressive}. Even though it is well-known that autoregressive decoding leads to higher accuracy scores, fast inference was one of our top priority features for \GT{}. 

\textbf{Flexible text editing.} 
While simplifying the learning task, text-editing models are not as powerful as general purpose sequence-to-sequence approaches when it comes to modeling arbitrary input-output text transductions. Hence, we strive to strike a balance between the complexity of learned edit operations and the percentage of input-output transformations the model can capture.

\paragraph{\textsc{Felix.}}
To meet the aforementioned desideratum, we propose to tackle text editing by decomposing it into two sub-problems: \tagging{} and \insertion{} (see Fig.~\ref{fig:gutentag}). Our \textit{tagger} is a Transformer-based network that implements a novel Pointing mechanism~\cite{PointerNetworks}. It decides which source tokens to preserve and in which order they appear in the output, thus allowing for arbitrary word re-ordering. 

The target words not present in the source are represented by the generic slot predictions to be in-filled by the \insertion{} model. To benefit from self-supervised pre-training, we chose our \insertion{} model to be fully compatible with the BERT architecture, such that we can easily re-use the publicly available pre-trained checkpoints.

By decomposing text-editing tasks in this way we redistribute the complexity load of generating an output text between the two models: the source text already provides most of the building blocks required to reconstruct the target, which is handled by the \textit{tagging} model. The missing pieces are then in-filled by the \insertion{} model, whose job becomes much easier as most of the output text is already in-place. Moreover, such a two-step approach is the key for being able to use completely non-autoregressive decoding for both models and still achieve competitive results compared to fully autoregressive approaches. 

We evaluate \GT{} on four distinct text generation tasks: Sentence Fusion, Text Simplification, Summarization, and Automatic Post-Editing for Machine Translation and compare it to recent text-editing and seq2seq approaches. 
Each task is unique in the editing operations required and the amount of training data available, which helps to better quantify the value of solutions we have integrated into \GT{}\footnote{The code is publicly available at: \textit{URL to be added}}.

\begin{figure*}[t]
\setlength{\tabcolsep}{3pt} 
\centering

\begin{tabular}{lllcccccc}
& \textbf{Src:} & The & big & very & loud & cat & \\
\midrule
\multirow{3}{*}{\textbf{Mask}} & \textbf{$\bv{y}^t$:} & \tags{\color{green}KEEP} & \tags{\color{red}DEL} & \tags{\color{red}DEL} & \tags{\color{red}DEL$^{\text{INS\_2}}$} & \tags{\color{green}KEEP} \\
& \textbf{$\bv{y}^m$:} & The & \texttt{[REPL]} big very loud \texttt{[/REPL]}  & \texttt{MASK} & \texttt{MASK} & cat & \\
& \textbf{Pred:} & The & & noisy & large & cat & \\
\midrule
\multirow{3}{*}{\textbf{Infill}} & \textbf{$\bv{y}^t$:} & \tags{\color{green}KEEP} & \tags{\color{red}DEL} & \tags{\color{red}DEL} & \tags{\color{red}DEL$^{\text{INS}}$} & \tags{\color{green}KEEP} \\
& \textbf{$\bv{y}^m$:} & The & \tags{[REPL]} big very loud \tags{[/REPL]} & \tags{MASK} & \tags{MASK} & \tags{MASK} & \tags{MASK} & cat \\
& \textbf{Pred:} & The & & noisy & large & \texttt{PAD} & \texttt{PAD} & cat \\
\end{tabular}
\caption{An example of two ways to model inputs to the \insertion{} model: via token \textit{masking} (Mask) or \textit{infilling} (Infill). In the former case the \tagging{} model predicts the number of masked tokens (\texttt{INS\_2}), while in the latter it is delegated to the \insertion{} model, which replaces the generic \texttt{INS} tag with a fixed length span (length 4). Note that the \insertion{} model predicts a special \texttt{PAD} symbol to mark the end of the predicted span. Replacements are modeled by keeping the deleted spans between the \tags{[REPL]} tags. This transforms the source text \textit{The big very loud cat} into the target \textit{The noisy large cat}. Note that for simplicity this example does not include reordering.}
\label{fig:masking_vs_infilling}
\end{figure*}

\section{Model description}
\GT{} decomposes the conditional probability of generating an output sequence $\bv{y}$ from an input ${\bv{x}}$ as follows: $p(\bv{y}|\bv{x}) = p_{\text{ins}}(\bv{y}|\bv{y}^{m})p_{\text{tag}}(\bv{y}^t,\pi|\bv{x})$, where the two terms correspond to the \tagging{} and the \insertion{} model. Term $\bv{y}^m$, which denotes an intermediate sequence with masked spans $\bv{y}^m$ fed into the \insertion{} model, is constructed from  $\bv{y}^t$, a sequence of tags assigned to each input token $\bv{x}$, and a permutation $\bv{\pi}$, which reorders the input tokens. Given this factorization, both models can be trained independently. 

\subsection{Tagging}

The tag sequence $\bv{y}^t$ is constructed as follows: source tokens that must be copied are assigned the \tags{KEEP} tag, tokens not present in the output are marked by the \tags{DELETE} tag, token spans present in the output but missing from the input are modeled by the \tags{INSERT} (\tags{INS}). This tag is then converted into masked token spans in-filled by the \insertion{} model. Word reordering is handled by a specialized Pointing mechanism.

\subsection{Pointing}

\begin{figure}
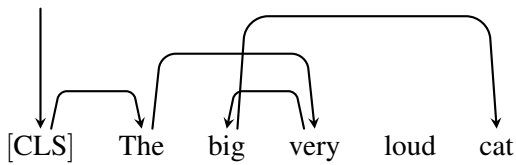

\centering
\begin{dependency}[hide label]
\begin{deptext}[column sep=.4cm, row sep=.2ex]
$[$CLS$]$ \& The \& big \& very \& loud \& cat \\
\end{deptext}
\deproot[edge style=thick]{1}{root}{}
\depedge[edge style=thick]{1}{2}{}
\depedge[edge style=thick]{2}{4}{}
\depedge[edge style=thick]{4}{3}{}
\depedge[edge style=thick]{3}{6}{}
\end{dependency}
\caption{Pointing mechanism to transform \textit{``the big very loud cat"} into \textit{``the very big cat"}.}
\label{fig:pointing_example}
\end{figure}

\GT{} explicitly models word reordering to allow for larger global edits, as well as smaller local changes, such as swapping nearby words, \emph{John and Mary} $\rightarrow$ \emph{Mary and John}.
Without word reordering step a vanilla editing model based on just tagging such as~\cite{LaserTagger,EditNTS}, would first need to delete a span (\emph{and Mary}) and then insert \emph{Mary and} before \emph{John}. \GT{} is able to model this without the need for deletions or insertions.

Given a sequence $\bv{x}$ and the predicted tags $\bv{y}^t$, the re-ordering model generates a permutation $\bv{\pi}$ so that from $\pi$ and $\bv{y}^t$ we can reconstruct the insertion model input $\bv{y}^m$. Thus we have: $p(\bv{y}^m | \bv{x}) = \prod_i p(\pi(i) | \bv{x}_i, \bv{y}^t) p(\bv{y}^t_i | \bv{x})$. 
We highlight that each $\pi(i)$ is  predicted independently in a non auto-regressive fashion. The output of this model is a series of predicted pointers (source token $\rightarrow$ next target token). $\bv{y}^m$ can easily be constructed by daisy chaining the pointers together, as seen in Fig.~\ref{fig:pointing_example}. As highlighted by this figure, \GT{}'s reordering process is similar to non-projective dependency parsing~\citet{DozatM17}, where head relationships are non-autoregressively predicted to form a tree. Similarly \GT{} predicts next word relationship and instead forms a sequence. 

Our implementation is based on a pointer network \citep{PointerNetworks}, where an attention mechanism points to the next token. Unlike previous approaches where a decoder state attends over an encoder sequence, our setup applies intra-attention, where source tokens attend to all other source tokens.

When constructing the training data there are many possible combinations of $\bv{\pi}$ and $\bv{y}^t$ which could produce $\bv{y}^m$, as trivially all source tokens can be deleted and then target tokens inserted. Hence, we construct the dataset using a greedy method to maximize the number of kept tokens, minimize the number of inserted token, and minimize the amount of reordering, keeping source tokens in continuous sequences where possible.  Since each token can only point to one other token, loops will be formed if the same token is pointed to multiple times. When constructing the dataset, we ensure that each token is only pointed to at most once. At inference time a constrained beam search is used to ensure no loops are created.

\subsection{Insertion}
An input to the \insertion{} model $\bv{y}^m$ contains a subset of the input tokens in the order determined by the \tagging{} model, as well as masked token spans that it needs to in-fill. 

To represent masked token spans we consider two options: \textit{masking} and \textit{infilling} (see Fig.~\ref{fig:masking_vs_infilling}). In the former case the \tagging{} model predicts how many tokens need to be inserted by specializing the \tags{INSERT} tag into \tags{INS\_k}, where \texttt{k} translates the span into \texttt{k} \texttt{MASK} tokens.

For the \textit{infilling} case the \tagging{} model predicts a generic \tags{INS} tag, which signals the \insertion{} model to infill it with a span of tokens of an arbitrary length. If we were to use an autoregressive \insertion{} model, the natural way to model it would be to run the decoder until it decides to stop by producing a special \textit{stop} symbol, e.g.,~\tags{EOS}. Since by design we opted for using a non-autoregressive model, to represent variable-length insertions we use a \tags{PAD} symbol to pad all insertions to a fixed-length sequence of \tags{MASK} tokens.\footnote{In all our experiments the maximum lengths of 8 was sufficient to represent over 99\% of insertion spans from the training set.}

Note that we preserve the deleted span in the input to the \insertion{} model by enclosing it between \tags{[REPL]} and \tags{[/REPL]} tags. Even though this introduces an undesired discrepancy between the pretraining and fine-tuning data that the \insertion{} model observes, we found that making the model aware of the text it needs to replace significantly boosts the accuracy of the \insertion{} model.

\subsection{\GT{} as Insertion Transformer}
Another intuitive way to picture how \GT{} works is to draw a connection with the Insertion Transformer~\cite{InsertionTransformer}. In the latter the decoder starts with a blank output text (canvas) and iteratively infills it by deciding which token and in which position it should appear in the output. Multiple tokens can be inserted at a time thus achieving sub-linear decoding times. In contrast, \GT{} trains a separate \tagger{} model to pre-fill\footnote{In the text edit tasks reported in this paper this corresponds to more than 80\% of the output tokens.} the output canvas with the input tokens in a single step. As the second and final step \GT{} does the insertion into the slots predicted by the \tagger{}. This is equivalent to a single decoding step of the Insertion Transformer. Hence, \GT{} requires significantly fewer (namely, two) decoding steps than Insertion Transformer, and through the \tagging{}/\insertion{} decomposition of the task it is straightforward to directly take advantage of existing pre-trained masked language models.

\section{Model implementation}

\subsection{Tagging Model}
\label{sec:LaserPointer}

\paragraph{Tagger.} Our \tagger{} is a 12-layer BERT-base model. Tags are predicted by applying a single feed-forward layer $f$ to the output of the encoder $\bv{h}^L$, as such $\bv{T} = \text{argmax}f(\bv{h}^L)$. 

\paragraph{Pointer.} The input to the Pointer layer at position $i$ is a combination of the encoder hidden state $\bv{h}_i^L$, the embedding of the predicted tag $e(\bv{T}_i)$ and the positional embedding $e(\bv{p}_{i})$\footnote{\citet{voita2019bottom} have shown that models trained with masked language modeling objectives  lose positional information, a property we consider important for reordering.} as follows: $\bv{h}^{L+1}_i = f([\bv{h}_i^L; e(\bv{T}_i); e(\bv{p}_{i})])$. 

Next token prediction uses a pointer network attending over all hidden states, as such:
\begin{equation}
p(\pi(i)|\bv{h}^{L+1}_i)= \text{attention}(\bv{h}^{L+1}_i, \bv{h}^{L+1}_{\pi(i)})
\end{equation}

Attention between hidden states is calculated using a query-key network with a scaled dot-product:
\begin{equation}
\text{Attention}(\bv{Q}, \bv{K}) = \text{softmax}(\frac{\bv{Q}\bv{K}^T}{\sqrt{d_k}})
\end{equation}

Where $\bv{K}$ and $\bv{Q}$ linear projections of $\bv{h}^{L+1}$ and $d_k$ is the hidden dimension. We found the optional inclusion of an additional Transformer layer prior to the query projection increased the performance on movement heavy datasets.

When realizing the pointers, we use a constrained beam search where we ensure no loops are created. We note that loops only form in $<3\%$ of the case\footnote{We fix the beam size to 5. For a batch size of 32 and maximum sequence length of 128, beam search incurs an additional penalty of about 12ms when run on a Xeon CPU@3.7GHz.}.

\subsection{Insertion Model}
Similar to the \tagger{}, our \insertion{} model is also based on a 12-layer BERT-base and is initialized from a public pretrained checkpoint. 

When using the \textit{masking} approach, the \insertion{} model is essentially  solving a masked language modeling task and, hence, we can directly take advantage of the BERT-style pretrained checkpoints. This is a considerable advantage especially in the low-resource settings as we do not waste training data on learning a language model component of the text-editing model. With the task decomposition where \tagging{} and \insertion{} can be trained disjointly it essentially comes for free\footnote{We still fine-tune the insertion model to accommodate for the additional token spans between the \tags{[REPL]} and \tags{[/REPL]}) such that it learns to condition the prediction of masked tokens on those spans.}.

Switching from \textit{masking} approach to \textit{infilling} shifts the complexity of modeling the length of the inserted token spans from the \tagging{} model to the \insertion{} model. Depending on the amount of training data available it provides interesting trade-offs between the accuracy of the \tagging{} and \insertion{} models. We explore this more in detail in Sec.~\ref{sec:discofuse}.

\begin{table*}[tb]
\centering

\begin{tabular}{lllllllll}
\toprule
Dataset & Size & $L_{\text{src}}$ & $L_{\text{tgt}}$ &TER & Ins & Del & Sub & Shft \\ 
\midrule
Post-editing & 5M & 18.10 & 17.74  & 24.97 & 04.24 & 06.25 & \textbf{11.30} & 02.69 \\ 

Simplification & 296K & 22.61& 21.65  & 26.02 & \textbf{04.75} & 08.97 & 09.90 & 02.41 \\ 
Summarization & \textbf{26K} & \textbf{32.48} & 22.16 & \textbf{43.23} & 00.29 & \textbf{32.06} & 09.34 & \textbf{10.71} \\ 
Sentence fusion & 4.5M & 30.51 & \textbf{30.04}  & 10.92 & 02.49 & 04.91 & 03.75 & 00.62 \\ 
\bottomrule
\end{tabular}
\caption{Statistics across tasks: size of the dataset (Size), source length in tokens ($L_{\text{src}}$), target length in tokens ($L_{\text{tgt}}$), and TER score~\cite{snover2006study} along with its components, including number of insertions (Ins), deletions (Del), substitutions (Sub), and shifts (Shft).}
\label{tbl:ter}
\end{table*}

\begin{table}[tb]
\centering
\resizebox{\columnwidth}{!}{%

\begin{tabular}{l|cc|cc}
\toprule
Dataset & \multicolumn{2}{c|}{Coverage} & \multicolumn{2}{c}{MASK \%} \\
 & \LT & \LP & \INSERTION & \GT\\ 
\midrule
Postediting & 35.10 & \textbf{40.40} &42.39 & \textbf{17.30} \\

Simplification & 36.87 & \textbf{42.27} & 18.23 & \textbf{13.85} \\
Summarization & 16.71 & \textbf{48.33} & 15.92 & \textbf{11.91} \\
Sentence fusion & 85.39 & \textbf{95.25} & 14.69 & \textbf{09.20} \\
\bottomrule
\end{tabular}
}
\caption{Coverage and MASK statistics. Coverage is the percentage of training examples that the models are able to generate. Both \INSERTION{} and \GT{} have full coverage of all test sets. MASK \% is the ratio of masked tokens to target tokens.}
\label{tbl:coverage}
\end{table}

\section{Experiments}
We evaluate \GT{} on four distinct text editing tasks: Sentence Fusion, Text Simplification, Summarization, and Automatic Post-Editing for Machine Translation. In addition to reporting previously published results for each task, we also compare to a recent text-editing approach \LT{}~\cite{LaserTagger}. We follow their setup and set the phrase vocabulary size to 500 and run all experiments using their most accurate autoregressive model. For all tasks we run an ablation study, examining the effect of an open vocabulary with no reordering (\INSERTION{}), and a fixed vocabulary\footnote{For simplicity we use the \LT{} phrase vocabulary.} with reordering model (\LP). 

\paragraph{Task analysis.} The chosen tasks cover a diverse set of edit operations and a wide range of dataset sizes, varying from under 30,000 data points to over 5 million. Table \ref{tbl:ter} provides dataset statistics including: the size, sentence length, and the translation error rate (TER)~\cite{snover2006study} between the source and target sentences. We use TER to highlight unique properties of each task. The summarization dataset is a deletion heavy dataset, with the highest number of deletion edits and the largest reduction in sentence length. It contains moderate amounts of substitutions and large number of shift edits, caused by sentence re-ordering. Both the simplification and post-editing datasets contain a large number of insertions and substitutions, while simplification contains a greater number of deletion edits.  Post-editing, however, is a much larger dataset covering multiple languages. Sentence fusion has the lowest TER, indicating that obtaining the fused targets requires only a limited number of local edits. However, these edits require modeling the discourse relation between the two input sentences, since a common edit type is predicting the correct discourse connective \cite{geva2019discofuse}.

Additionally, we provide coverage statistics and the percentage of training instances for which an editing model can fully reconstruct the output from the input of our proposed model in Table~\ref{tbl:coverage}, contrasting it against \LT{}. As both \GT{} and \INSERTION{} use an open vocabulary, they cover 100\% of the test data, whereas \LP{} and \LT{} often cover less than half. For every dataset \LP{} covers a significantly higher percentage than \LT{}, with the noticeable case being summarization, where there is a 3x  increase in coverage. This can be explained by the high number of shift edits within summarization (Table~\ref{tbl:ter}), something \LP{} is explicitly designed to model. We found that the difference in coverage between \LP{} and \LT{} correlates strongly (correlation 0.99, p$<$0.001) with the number of shift edits. Comparing the average percentage of MASKs inserted, we see that \GT{} always inserts ($\sim$50\%) less MASKs than \INSERTION, since no word reordering requires more deletions and insertions for the latter.

\begin{table*}[tb]
\footnotesize
\centering
\begin{tabular}{l|cc|cc|llllll}
\toprule
Model & \multicolumn{2}{c|}{Insertion} & \multicolumn{2}{c|}{Oracle} & SARI & Exact & 10\% & 1\% & 0.1\% & 0.01\%  \\     
& Mask & Infill & TAG & INS & & & & & & \\     
\midrule
\bertobert & & & & & \textbf{89.52} & \textbf{63.90} & 54.45 & 42.07 & 03.35 & 00.00 \\
\seqtoseqbert & & & & & 85.30 & 53.60 & 52.80 & 43.70 & 00.00 & 00.00 \\
\LT{} & & & & & 85.45 & 53.80 &  47.31 & 38.46& 25.74 & 12.32\\ 
\midrule
\LP & & & & & 88.20 & 60.76 & 53.75 & 44.90 & 31.87 & 13.82 \\ 
\midrule

\multirow{6}{*}{\INSERTION} & \cellcolor{gray!50} & \cellcolor{gray!50}$\bullet$ &  \cellcolor{gray!50}$\bullet$ &\cellcolor{gray!50} &\cellcolor{gray!50} & \cellcolor{gray!50}82.91 & \cellcolor{gray!50}77.25 & \cellcolor{gray!50}71.49 & \cellcolor{gray!50}57.94 & \cellcolor{gray!50}36.61  \\
& \cellcolor{gray!30} &  \cellcolor{gray!30}$\bullet$ & \cellcolor{gray!30}&  \cellcolor{gray!30}$\bullet$ &\cellcolor{gray!30} & \cellcolor{gray!30}75.00 & \cellcolor{gray!30}71.97 & \cellcolor{gray!30}66.87 & \cellcolor{gray!30}57.08 & \cellcolor{gray!30}38.89 \\
& &  $\bullet$ & & & 88.44& 60.80 & 52.82 & 46.09 & 34.11 & 15.34  \\
& \cellcolor{gray!50}$\bullet$ &\cellcolor{gray!50} &  \cellcolor{gray!50}$\bullet$ &\cellcolor{gray!50} & \cellcolor{gray!50}& \cellcolor{gray!50}72.91&  \cellcolor{gray!50}64.00 & \cellcolor{gray!50}55.45 & \cellcolor{gray!50}39.71 & \cellcolor{gray!50}18.89 \\
& \cellcolor{gray!30}$\bullet$ & \cellcolor{gray!30} &\cellcolor{gray!30}  &  \cellcolor{gray!30}$\bullet$ & \cellcolor{gray!30}& \cellcolor{gray!30}88.86 & \cellcolor{gray!30}84.11 & \cellcolor{gray!30}81.76 & \cellcolor{gray!30}75.88 & \cellcolor{gray!30}61.68 \\ 
&  $\bullet$ & & & & 88.72 & 63.37 & \textbf{56.67} & \textbf{48.85} & 33.32 & 13.99  \\
\midrule

\multirow{6}{*}{\textbf{\GT}} & \cellcolor{gray!50} &  \cellcolor{gray!50}$\bullet$ &  \cellcolor{gray!50}$\bullet$ &\cellcolor{gray!50} & \cellcolor{gray!50}& \cellcolor{gray!50}70.32 & \cellcolor{gray!50}71.78 & \cellcolor{gray!50}64.28 & \cellcolor{gray!50}51.20 & \cellcolor{gray!50}28.42\\
& \cellcolor{gray!30} & \cellcolor{gray!30}$\bullet$ &\cellcolor{gray!30} & \cellcolor{gray!30}$\bullet$ & \cellcolor{gray!30}& \cellcolor{gray!30}78.37 & \cellcolor{gray!30}75.56 & \cellcolor{gray!30}72.24 & \cellcolor{gray!30}65.95 & \cellcolor{gray!30}55.97\\
& & $\bullet$ & & & 87.69 & 58.32 & 55.11 & 48.84 & \textbf{38.01} & \textbf{20.49} \\
& \cellcolor{gray!50}$\bullet$ &\cellcolor{gray!50}  &\cellcolor{gray!50}$\bullet$ &\cellcolor{gray!50}  &\cellcolor{gray!50}   &\cellcolor{gray!50}67.78 &\cellcolor{gray!50}59.62 &\cellcolor{gray!50}52.74 &\cellcolor{gray!50}41.48 &\cellcolor{gray!50}17.30 \\  
& \cellcolor{gray!30}$\bullet$ &\cellcolor{gray!30}  &\cellcolor{gray!30}  &\cellcolor{gray!30}$\bullet$ &\cellcolor{gray!30}  & \cellcolor{gray!30}87.52 &\cellcolor{gray!30}86.45 &\cellcolor{gray!30}83.13 &\cellcolor{gray!30}79.79 &\cellcolor{gray!30}67.60 \\  
& $\bullet$ & & & & 88.78 & 61.31 & 52.85 & 45.45 & 36.87 & 16.96 \\

\bottomrule
\end{tabular}
\caption{Sentence Fusion results on DiscoFuse using the full and subsets 10\%, 1\%, 0.1\% and 0.01\% of the training set. We report three model variants: \LP{}, \INSERTION{} and \GT{} using either Mask or Infill insertion modes. Rows in gray background report scores assuming oracle \tagging{} (TAG) or \insertion{} (INS) predictions.}
\label{tbl:dfwiki}
\end{table*}

\subsection{Sentence Fusion}
\label{sec:discofuse}
Sentence Fusion is the problem of fusing independent sentences into a coherent output sentence(s).

\paragraph{Data.} We use the “balanced Wikipedia” portion of the DiscoFuse dataset~\cite{geva2019discofuse} and also study the effect of the training data size by creating four increasingly smaller subsets of DiscoFuse: 450,000 (10\%), 45,000 (1\%), 4,500 (0.1\%) and 450 (0.01\%) data points. 

\paragraph{Metrics.} Following \citet{geva2019discofuse}, we report two metrics: \textit{Exact score}, which is the percentage of exactly correctly predicted fusions, and \textit{SARI} \citep{xu-etal-2016-optimizing}, which computes the average F1 scores of the added, kept, and deleted n-grams.

\paragraph{Results.} Table~\ref{tbl:dfwiki} includes additional BERT-based seq2seq baselines: \bertobert{} from~\cite{Bert2Bert} and \seqtoseqbert{} from~\cite{LaserTagger}. For all \GT{} variants we further break down the scores based on how the \tags{INSERTION} is modelled: via token-masking (Mask) or Infilling (Infill). Additionally, to better understand the contribution of \tagging{} and \insertion{} models to the final accuracy, we report scores assuming oracle \insertion{} and \tagging{} predictions respectively (highlighted rows). 

The results show that \GT{}  and its variants significantly outperform the baselines \LT{} and \seqtoseqbert{}, across all data conditions. Under the 100\% condition  \bertobert{} achieves the highest SARI and Exact score, however for all other data conditions \GT{} outperforms \bertobert. The results highlights that both  seq2seq models perform poorly with less than 4500 (0.1\%) datapoints, whereas all editing models achieve relatively good performance.

When comparing \GT{} variants we see that in the 100\% case \INSERTION{} outperforms \GT{}, however we note that for \INSERTION{} we followed \citep{LaserTagger} and used an additional sentence re-ordering tag, a hand crafted feature tailored to DiscoFuse which swaps the sentence order. It was included in \citep{LaserTagger} and resulted in a significant (6\% Exact) increase. However, in the low resource setting  \GT{} outperforms \INSERTION{}, suggesting that \GT{} is more data efficient than \INSERTION.

\paragraph{Ablation.}
We first contrast the impact of the \insertion{} model and the \tagging{} model, noticing that for all models Infill achieves better tagging scores and worse insertion scores than Mask. Secondly, \GT{} achieves worse tagging scores but better insertion scores than \INSERTION. This highlights the amount of pressure each model is doing, by making the tagging task harder, such as the inclusion of reordering, the insertion task becomes easier. Finally, the \insertion{} models even under very low data conditions achieve impressive performance. This suggests that under low data conditions most pressure should be applied to the \insertion{} model.

\subsection{Simplification}
Sentence simplification  is the problem of simplifying sentences such that they are easier to understand. Simplification can be both lexical, replacing or deleting complex words, or syntactic, replacing complex syntactic constructions.

\paragraph{Data.} Training is performed on WikiLarge, \cite{zhang2017sentence} a large simplification corpus which consists of a mixture of of three Wikipedia simplification datasets collected by \cite{kauchak2013improving,woodsend2011learning,zhu2010monolingual}. The test set  was created by \citet{xu-etal-2016-optimizing} and consists of 359 source sentences taken from Wikipedia, and then simplified using Amazon Mechanical Turkers to create eight references per source sentence. 

\paragraph{Metrics.} We report SARI, as well as breaking it down into each component \tags{KEEP}, \tags{DELETE}, and \tags{ADD}, as we found the scores were uneven across these metrics. We include a readability metrics (FKGL), and the percentage of unchanged source sentences (copy). 

\paragraph{Results.} In Table \ref{tbl:wikilarge} we compare against three state-of-the-art SMT based simplification systems: (1) PBMT-R \cite{wubben-etal-2012-sentence},  a phrase-based machine translation model. (2) Hybrid \cite{narayan2014hybrid}, a model which performs sentence spiting and deletions and then simplifies with PBMT-R.  (3) SBMT-SARI \cite{xu-etal-2016-optimizing}, a syntax-based translation model trained on PPDB and which is then tuned using SARI. Four neural seq2seq approaches: (1) DRESS \cite{D17-1063}, an  LSTM-based seq2seq trained with reinforcement learning (2) DRESS-Ls,  a variant of DRESS which has an additional lexical simplification component; (3) NTS \cite{nisioi2017exploring} and seq2seq model. (4) DMASS \cite{zhao2018integrating}, a transformer-based model enhanced with simplification rules from PPDB. And two neural editing models: (1) \LT{} and (2) EditNTS. 

\GT{} achieves the highest overall SARI score and the highest SARI-KEEP score. In addition, all ablated models achieve higher SARI scores than \LT{}. While \INSERTION{} achieves a higher SARI score than EditNTS \LP{} does not, this can in part be explained by the large number of substitutions and insertions within this dataset, with \LP{} achieving a low SARI-ADD score.

\begin{table}[tb]
\centering
\resizebox{\columnwidth}{!}{%
\footnotesize
\begin{tabular}{p{1.9cm}p{0.55cm}p{0.55cm}p{0.55cm}p{0.55cm}p{0.55cm}p{0.55cm}}
\toprule
 WikiLarge & SARI & ADD & DEL & KEEP & FKGL & Copy \\
\midrule
{\sc SBMT-SARI} &37.94 & \textbf{05.60} & 37.96 & 70.27 & 8.89 & 0.10 \\
{\sc DMASS+DCSS} & 37.01 & 05.16 & 40.90 & 64.96 & 9.24 & 0.06 \\
{\sc PBMT-R} & 35.92 & 05.44 & 32.07 & 70.26 & 10.16 & 0.11 \\
{\sc Hybrid} & 28.75 & 01.38 & \textbf{41.45} & 43.42 & \textbf{7.85} & \textbf{0.04} \\
{\sc NTS} & 33.97 & 03.57 & 30.02 & 68.31 & 9.63 & 0.11 \\
{\sc DRESS} & 33.30 & 02.74 & 32.93 & 64.23 & 8.79 & 0.22 \\
{\sc DRESS-LS} & 32.98 & 02.57 & 30.77 & 65.60 & 8.94 & 0.27 \\
{\sc EditNTS} & 34.94 & 03.23 & 32.37 & 69.22 & 9.42 & 0.12 \\
\LT & 32.31 & 03.02 & 33.63 & 60.27 & 9.82 & 0.21 \\ 
\midrule 
\LP & 34.37 & 02.35 & 34.80 & 65.97 & 9.47 & 0.18 \\
\INSERTION{} & 35.79 & 04.03 & 39.70 & 63.64 & 8.14 & 0.09 \\
\GT & \textbf{38.13} & 03.55 & 40.45 & \textbf{70.39} & 8.98 & 0.08 \\
\bottomrule
\end{tabular}
}
\caption{Sentence Simplification results on WikiLarge. }
\label{tbl:wikilarge}
\end{table}

\subsection{Summarization}
\paragraph{Data.}
We use the dataset from \cite{toutanova2016dataset}, which contains 6,168 short input texts (one or two sentences) and one or more human-written summaries, resulting in 26,000 total training pairs. The human experts were not restricted to just deleting words when generating a summary, but were allowed to also insert new words and reorder parts of the sentence, which makes this dataset particularly suited for abstractive summarization models.

\paragraph{Metrics.}
 In addition to SARI we include ROUGE-L and BLEU-4, as these  metrics are commonly used in the summarization literature. 

\paragraph{Results.} 
The results in Table \ref{tbl:sum} show that \GT{} achieves the highest SARI, ROUGE and BLEU score. All ablated models achieve higher SARI scores than all other models. Interestingly, the difference between \LP{} and \LT{} is modest, even though \LP{} covers twice as much data as \LT. With \LT{} being trained on ~4500 data points and \LP{}  trained on 13000. In Table~\ref{tbl:dfwiki} we see that \LT{} and \LP{} perform similarly under such low data conditions.

\begin{table}[tb]
\centering
\resizebox{\columnwidth}{!}{%
\footnotesize
\begin{tabular}{p{2cm}p{0.55cm}p{0.55cm}p{0.55cm}p{0.55cm}p{0.55cm}p{0.55cm}}
\toprule
& SARI & ADD & DEL & KEEP & Rouge & BLEU \\ 
\midrule
\seqtoseqbert & 32.10 & & & & 52.70 & 08.30 \\
\LT & 40.36 & 06.04 & 54.47 & 60.57 & 81.68 & 35.47 \\
\midrule
\LP & 40.97 & 05.94 & 58.30 & 58.67 & 79.47 & 31.34 \\
\INSERTION{} & 41.85 & 06.45 & \textbf{61.37} & 57.73 & 78.12 & 29.78 \\
\GT & \textbf{42.60} & \textbf{07.65} & 57.26 & \textbf{62.89} & \textbf{83.54} & \textbf{36.23} \\
\bottomrule
\end{tabular}
}
\caption{Summarization.  Copy is not included as all models copied less than 2\% of the time.}
\label{tbl:sum}
\end{table}

\subsection{Post-Editing}
Automatic Post-Editing (APE) is the task of automatically correcting common and repetitive errors found in machine translation (MT) outputs. 

\paragraph{Data.}
APE approaches are trained on triples: the source sentence, the machine translation output, and the target translation. We experiment on the WMT17 EN-DE IT post-editing task\footnote{\url{http://statmt.org/wmt17/ape-task.html}}, where the goal is to improve the output of an MT system that translates from English to German and is applied to documents from the IT domain. We follow the procedures introduced in \citep{junczysdowmunt-grundkiewicz:2016:WMT} and train our models using two synthetic corpora of 4M and 500K examples merged with a corpus of 11K real examples over-sampled 10 times. The models that we study expect a single input string. To obtain this and to give the models a possibility to attend to the English source text, we append the source text to the German translation separated by a special token. Since the model input consists of two different languages, we use the multilingual BERT checkpoint\footnote{\url{https://storage.googleapis.com/bert_models/2018_11_23/multi_cased_L-12_H-768_A-12.zip}} for the proposed methods and for \LT{}.

\paragraph{Metrics.} We follow the evaluation procedure of WMT17 APE task and report translation error rate (TER) \cite{snover2006study} as the primary metric and BLEU as a secondary metric.

\paragraph{Results.} We consider the following baselines: {\sc Copy}, which is a competitive baseline given that the required edits are typically very limited, \LT{} \cite{LaserTagger}, {\sc Levenshtein Transformer} (\LevT) \cite{LevenshteinTransformer}, which is a partially autoregressive model that also employs a \textit{deletion} and an \textit{insertion} mechanisms, a standard {\sc Transformer} evaluated by \citep{LevenshteinTransformer}, and a state-of-the-art method by \cite{lee2019transformer}. Unlike the other methods, the last baseline is tailored specifically for the APE task by encoding the source separately and conditioning the MT output encoding on the source encoding \cite{lee2019transformer}.

The results are shown in Table~\ref{tbl:ape}. First, we can see that using a custom method \cite{lee2019transformer} brings significant improvements over generic text transduction methods. Second, \GT{} performs very competitively, yielding comparative results to {\sc Levenshtein Transformer} \cite{LevenshteinTransformer} which is a partially autoregressive model, and outperforming the other generic models in terms of TER. Third, \INSERTION{} performs considerably worse than \GT{} and \LP{}, suggesting that the pointing mechanism is important for the APE task. This observation is further backed by Table~\ref{tbl:coverage} which shows that without the pointing mechanism the average proportion of masked tokens in a target is 42.39\% whereas with pointing it is only 17.30\%. Therefore, removing the pointing mechanism shifts the responsibility too heavily from the tagging model to the insertion model.

\begin{table}[tb]
\centering
\footnotesize
\begin{tabular}{lll}
\toprule
& TER $\downarrow$ & BLEU $\uparrow$ \\
\midrule
{\sc Copy} & 24.48	 &62.49 \\
{\sc Transformer} & 22.1 & 67.2 \\ 
\LT & 24.29 & 63.83 \\
\LevT & 21.9 & 66.9 \\
SOTA \cite{lee2019transformer} & \textbf{18.13} & \textbf{71.80} \\ 
\midrule
\LP & 22.51 & 65.61 \\
\INSERTION{} & 29.09 & 57.42 \\
\GT & 21.87 & 66.74 \\
\bottomrule
\end{tabular}
\caption{WMT17 En$\rightarrow$De post-editing results.}
\label{tbl:ape}
\end{table}

\section{Related work}

Seq2seq models~\cite{sutskever2014sequence} have been applied to many text generation tasks that can be cast as monolingual translation, but they suffer from well-known drawbacks~\cite{wiseman2018learning}: they require large amounts of training data, and their outputs are difficult to control. Whenever input and output sequences have a large overlap, it is reasonable to cast the problem as a text editing task, rather than full-fledged sequence to sequence generation. \citet{ribeiro2018local}  argued that the general problem of string transduction can be reduced to sequence labeling. Their approach applied only to character deletion and insertion and was based on simple patterns. LaserTagger~\cite{LaserTagger} is a general approach that has been shown to perform well on a number of text editing tasks, but it has two limitations: it does not allow for arbitrary reordering of the input tokens; and insertions are restricted to a fixed phrase vocabulary that is derived from the training data. Similarly, EditNTS~\cite{EditNTS} and PIE~\cite{awasthi2019parallel} are two other text-editing models that predict tokens to keep, delete, and add, which are developed specifically for the tasks of text simplification and grammatical error correction, respectively. In contrast to the aforementioned models, \GT{} allows more flexible rewriting, using a pointer network that points into the source to decide which tokens should be preserved in the output and in which order.

Pointer networks have been previously proposed as a way to copy parts of
the input in hybrid sequence-to-sequence models. \citet{gulcehre2016pointing} and
\citet{nallapati2016abstractive} trained a pointer network to specifically deal with out-of-vocabulary
words or named entities.  \citet{see2017get} hybrid approach learns when to use the pointer
to copy parts of the input. \citet{chen2018fast} proposed a summarization model that first
selects salient sentences and then rewrites them abstractively, using a pointer mechanism to
directly copy some out-of-vocabulary words. These methods still typically require large amounts of training data and they are inherently slow at inference time due to autoregressive decoding.

Previous approaches have proposed
alternatives to autoregressive decoding~\cite{gu2018nonautoregressive,lee2018deterministic,KERMIT,wang2019bert}.
Instead of the left-to-right autoregressive decoding,
Insertion Transformer~\cite{InsertionTransformer} and BLM~\cite{shen2020blank} generate the output sequence
through insertion operations, whereas Levenshtein Transformer (\LevT{})~\cite{LevenshteinTransformer} additionally incorporates a deletion operation.

These methods produce the output iteratively, while \GT{} requires only two steps: \tagging{} and \insertion{}.

\begin{table}[tb]
\centering
\resizebox{\columnwidth}{!}{%

\begingroup
\setlength{\tabcolsep}{3pt} 
\renewcommand{\arraystretch}{1} 

\begin{tabular}{lccccc}
\toprule
& Type & \begin{tabular}[c]{@{}c@{}}Non-autore-\\gressive\end{tabular} & Pretrained & Reordering & \begin{tabular}[c]{@{}c@{}}Open\\vocab\end{tabular} \\
\midrule
{\sc Transformer} & \multirow{3}{*}{seq2seq} &  &  &  & \checkmark \\
$+$ {\sc Copying} & &  &  & \checkmark & \checkmark \\
{\sc T5} &  &  & \checkmark & (\checkmark) & \checkmark  \\
\midrule
\LevT & \multirow{7}{*}{Text edit} & (\checkmark) & \checkmark & & \checkmark  \\
{\sc PIE} & &  & \checkmark & & \checkmark  \\
{\sc EditNTS} & & & & & \checkmark  \\
\LT & & \checkmark & \checkmark & &  \\
\midrule
\INSERTION & & \checkmark & \checkmark & & \checkmark \\
\LP & & \checkmark & \checkmark & \checkmark & \\
\GT & & \checkmark & \checkmark & \checkmark  & \checkmark \\
\bottomrule
\end{tabular}
\endgroup
}
\caption{Model comparison along five dimensions: model \textit{type}, whether the decoder is \textit{non-autoregressive} (\LevT{} is partially autoregressive), whether the model uses a \textit{pretrained} checkpoint, a word \textit{reordering} mechanism ({\sc T5} uses a reordering pretraining task but it does not have a dedicated copying mechanism for performing reordering), and whether the model can generate any possible output (\textit{Open vocab}).}
\label{tbl:comparison}
\end{table}

The differences between the proposed model, \GT{}, its ablated variants, and a selection of related works is summarized in Table~\ref{tbl:comparison}.

\section{Conclusions and Future Work}
We have introduced \GT{}, a novel 
approach to text editing, by decomposing the task into \tagging{} and \insertion{} which are trained independently. Such separation allows us to take maximal benefit from the already existing pretrained masked-LM models. \GT{} works extremely well in low-resource settings and it is fully non-autoregressive which favors faster inference. Our empirical results demonstrate that it delivers highly competitive performance when compared to strong seq2seq baselines and other recent text editing approaches. 

In the future work we plan to investigate the following ideas: (i) how to effectively share represenations between the \tagging{} and \insertion{} models using a single shared encoder, (ii) how to perform joint training of \insertion{} and \tagging{} models instead of training them separately, (iii) strategies for unsupervised pre-training of the \tagging{} model which appears to be the bottleneck in highly low-resource settings, and (iv) distillations recipes.

\section*{Acknowledgments}
We thank Aleksandr Chuklin, Daniil Mirylenka, Ryan McDonald, and Sebastian Krause for useful discussions, running early experiments and paper suggestions.

\bibliography{main.bib}

\bibliographystyle{acl_natbib.bst}

\appendix

\end{document}